\documentclass[10pt,twocolumn,letterpaper]{article}

\usepackage{titling}
\usepackage{iccv}
\usepackage{subcaption}
\usepackage{times}
\usepackage{epsfig}
\usepackage{graphicx}
\usepackage{amsmath}
\usepackage{amssymb}
\usepackage{lipsum}  
\usepackage{graphicx}
\usepackage{amsmath}
\usepackage{amssymb}
\usepackage{booktabs}

\usepackage[utf8]{inputenc} 
\usepackage[T1]{fontenc}    
\usepackage{url}            
\usepackage{booktabs}       
\usepackage{amsfonts}       
\usepackage{nicefrac}       
\usepackage{microtype}      

\usepackage{multirow}
\usepackage{xspace}
\usepackage{mathtools}
\usepackage{wasysym}
\usepackage{marvosym}
\usepackage{xcolor}
\usepackage{color}
\usepackage{tabularx}
\usepackage{float}
\usepackage{diagbox,tabu,stackengine}
\usepackage{bbm}
\usepackage{bm}
\usepackage{mwe}
\usepackage{graphbox}
\usepackage{sidecap}
\usepackage{gensymb}
\usepackage{tabularray}
\usepackage{algorithm}
\usepackage[noend]{algpseudocode}
\usepackage{comment}
\usepackage{cancel}
\usepackage{ulem}
\usepackage{array}
\newcommand{\PreserveBackslash}[1]{\let\temp=\\#1\let\\=\temp}
\newcolumntype{C}[1]{>{\PreserveBackslash\centering}p{#1}}
\newcolumntype{R}[1]{>{\PreserveBackslash\raggedleft}p{#1}}
\newcolumntype{L}[1]{>{\PreserveBackslash\raggedright}p{#1}}

\usepackage{xspace}
\makeatletter
\DeclareRobustCommand\onedot{\futurelet\@let@token\@onedot}
\def\@onedot{\ifx\@let@token.\else.\null\fi\xspace}

\makeatother

\usepackage[pagebackref=true,breaklinks=true,letterpaper=true,colorlinks,bookmarks=false]{hyperref}

\iccvfinalcopy 

\def\iccvPaperID{****} 

\ificcvfinal\fi

\begin{document}

\title{Augmented Reality based Simulated Data (ARSim) with multi-view consistency for AV perception networks}

\author{Aqeel Anwar\\
NVIDIA\\
{\tt\small manwar@nvidia.com}
\and
Tae Eun Choe\\
NVIDIA\\
{\tt\small tchoe@nvidia.com}
\and
Zian Wang\\
NVIDIA / University of Toronto\\
{\tt\small zianw@nvidia.com}
\and
Sanja Fidler\\
NVIDIA / University of Toronto\\
{\tt\small sfidler@nvidia.com}
\and
Minwoo Park\\
NVIDIA\\
{\tt\small minwoop@nvidia.com}
}

\maketitle
\pagestyle{plain}
\ificcvfinal\thispagestyle{empty}\fi

\begin{abstract}
Detecting a diverse range of objects under various driving scenarios is essential for the effectiveness of autonomous driving systems. However, the real-world data collected often lacks the necessary diversity presenting a long-tail distribution. Although synthetic data has been utilized to overcome this issue by generating virtual scenes, it faces hurdles such as a significant domain gap and the substantial efforts required from 3D artists to create realistic environments. To overcome these challenges, we present ARSim, a fully automated, comprehensive, modular framework designed to enhance real multi-view image data with 3D synthetic objects of interest. The proposed method integrates domain adaptation and randomization strategies to address covariate shift between real and simulated data by inferring essential domain attributes from real data and employing simulation-based randomization for other attributes. We construct a simplified virtual scene using real data and strategically place 3D synthetic assets within it. Illumination is achieved by estimating light distribution from multiple images capturing the surroundings of the vehicle. Camera parameters from real data are employed to render synthetic assets in each frame. The resulting augmented multi-view consistent dataset is used to train a multi-camera perception network for autonomous vehicles. Experimental results on various AV perception tasks demonstrate the superior performance of networks trained on the augmented dataset.
\end{abstract}


\begin{figure*}
\centering
\includegraphics[width=\linewidth]{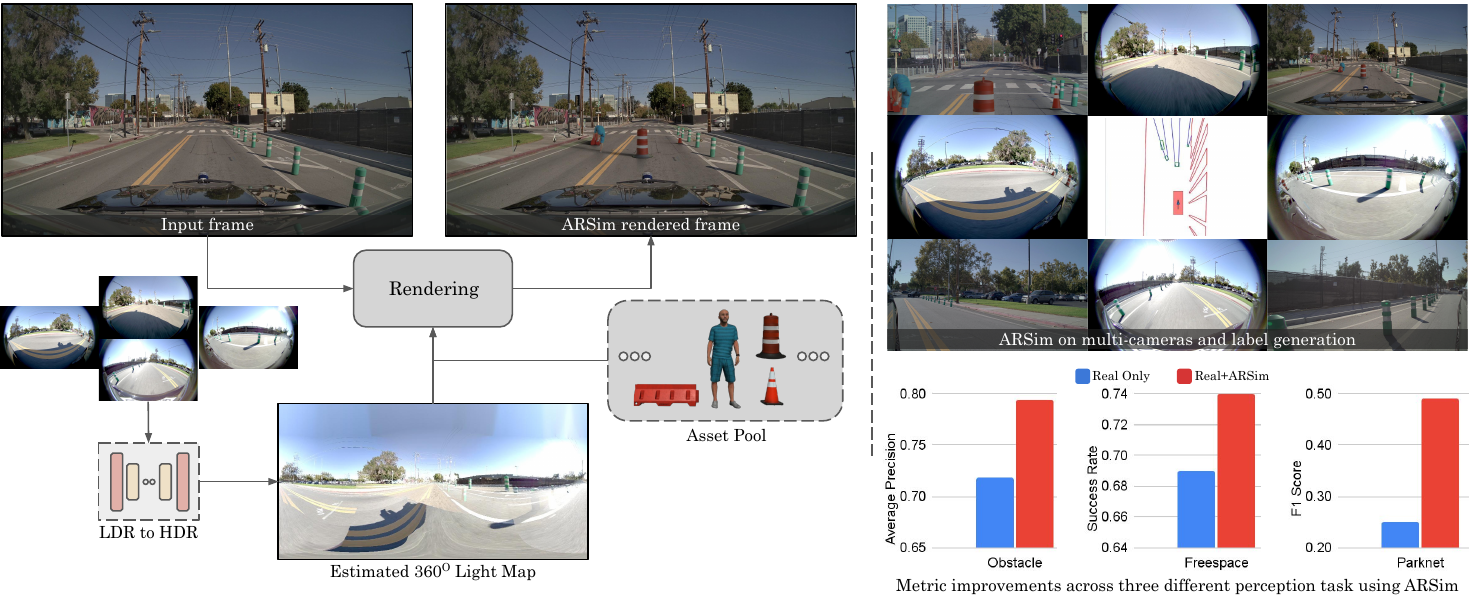}
\caption{An overview of the proposed approach and its impact. (left) We generate an HDR light map from the input data and position assets of interest in 3D around the ego car, subsequently rendering them within the camera frame, handling collision and occlusion with existing real objects. Concurrently, the pipeline achieves multi-view consistent frame rendering (right top). Additionally, integrating ARSim data with real data enhances performance metrics across three crucial AV perception tasks: obstacle detection, freespace detection, and parking detection (right bottom), as demonstrated in detail in the results section.}
\label{fig:feature}
\end{figure*}

\section{Introduction}
\label{sec:intro}

Autonomous vehicles (AV) heavily depend on perception systems to navigate safely and accurately. Key to this navigation is the ability to perceive obstacles, enabling vehicles to detect and steer clear of potential collisions on the road. In recent years, there has been a notable trend towards the development of 3D obstacle detection systems, leveraging the power of machine learning and camera sensors exclusively \cite{philion2020lift, Reading_2021_CVPR, Wang_2021_ICCV, li2022bevformer}. Despite these advancements, a key challenge in perception remains the detection of rarely-seen objects/scenarios, essential for ensuring road safety and preventing accidents. Robust perception models necessitate substantial-high-quality training data, especially for rarely observed and potentially hazardous events like traffic hazards, pedestrians, children, bicycles, and motorbikes. The scarcity of data for such scenarios poses a significant hurdle in creating accurate and reliable detection systems. Motivated by the limitations of real-world data, including scarcity and difficulties in obtaining diverse, annotated datasets, researchers explore alternative approaches. Additionally, the associated cost and time constraints in collecting real-world data have led to the adoption of synthetic data generation as a more efficient and cost-effective solution.

Synthetic data generation using 3D rendering platforms can be used to replicate various traffic scenarios with impeccable labels. However, a considerable domain gap between synthetic and real-world data can lead to sub-optimal performance in resulting models. While computer graphics (CG) artists can manually craft high-quality photorealistic scenes, this process is often expensive and lacks scalability. Another approach is to use the real-drive datasets and insert synthetic assets in the camera frames. Proposed copy-paste methods \cite{Choe_2023_CVPR, Ghiasi_2021_CVPR, pmlr-v202-zhao23f, Dwibedi_2017_ICCV, Dvornik_2018_ECCV} attempt to address data scarcity by cropping and pasting existing segmented 2D objects into new ones to simulate different scenarios. However, these methods fall short in providing multi-view consistent image editing and 3D groundtruth labels, essential for training multi-view 3D perception models.


In this paper, we present a novel method for automatically generating large-scale augmented data tailored for rarely captured objects. Our approach draws inspiration from lighting estimation and virtual object insertion techniques employed in the visual effects and augmented reality industries. The objective is to seamlessly and convincingly composite virtual objects into real-world images. We adapt and re-purpose this image editing technique to generate simulated objects of interest. This method aims to address the challenges posed by scarce real-world data, enhancing the training process and contributing to the accuracy and reliability of perception models in autonomous vehicles. 
Our contribution is as follows
\begin{itemize}
    \item We introduce ARSim, a fully automated, end-to-end framework designed to generate high-quality synthetic data by seamlessly enhancing real drive datasets with synthetic objects, thereby reducing the domain gap.
    \item We demonstrate the framework's capability to maintain high-quality realism and multi-view consistency through accurate inference of light distribution, proper shadow casting, and consistent rendering using camera parameters across all frames.
    \item We utilize ARSim to enrich real drive datasets with a focus on long-tail distribution objects and showcase the augmented dataset's efficacy in improving performance across three distinct AV tasks — obstacle perception, freespace perception, and parking perception — using only a fraction of the ARSim-generated data.
\end{itemize}

The paper is structured as follows. We begin by discussing related work in section \ref{sec:related-work} and draw comparisons with our proposed approach. In Section \ref{sec:problem-formulation}, we define the problem and introduce our proposed approach. Section \ref{sec:method} presents the modular ARSim pipeline, emphasizing the functions and applications of its components. Section \ref{sec:results} offers an examination of ARSim data generation for three distinct autonomous vehicle tasks, along with a thorough analysis of the quantitative enhancements resulting from the incorporation of ARSim data. In section \ref{sec:ablation} we carry out a comparative analysis of quantitative improvement for the task of 3D obstacle detection using purely synthetic data and the proposed ARSim data.

\section{Related Work}
\label{sec:related-work}

\subsubsection*{Synthetic Data Generation}
The domain of synthetic data generation for autonomous vehicles has experienced considerable growth, marked by the development of various simulators, including DriveSim \cite{DRIVESim}, CARLA \cite{dosovitskiy2017carla}, LGSVL \cite{rong2020lgsvl}, rFPro, TORCS \cite{wymann2000torcs}, Webots, and CoppeliaSim. These simulators play a crucial role in supporting the development, training, and validation of autonomous driving systems. They exhibit diversity in their focus, features, and applications, offering varying degrees of realism, customization, and sensor support.
Despite the sophistication of these platforms, which provides an enhanced degree of realism, there remains a noticeable distribution gap when compared to real-world datasets.
In complement to these simulators, machine learning-based approaches have emerged as a means of generating synthetic data. Generative Adversarial Networks (GANs) \cite{radford2015unsupervised, brock2018large, karras2019style, shaham2019singan} have gained widespread adoption for this purpose, involving the training of a generator to produce realistic samples and a discriminator to differentiate between real and synthetic data. While GANs contribute to improving the diversity and quality of generated data, challenges persist in ensuring multi-view consistency and generating necessary ground truth.

\subsubsection*{Domain Adaptation}
Efforts to mitigate the drawbacks of purely synthetic data have brought about the prominence of synthetic-to-real domain adaptation techniques. Techniques such as image-image translation methods \cite{zhu2017unpaired, isola2017image} and Domain Adversarial Neural Network (DANN) \cite{JMLR:v17:15-239} aim to transfer knowledge from synthetic to real data domains, thereby reducing the domain gap and improving model generalization. Additional approaches include metric learning \cite{kulis2011you, gong2012geodesic}, deep feature alignment \cite{tzeng2014deep}, and domain stylization \cite{dundar2018domain}.
\subsubsection*{Image Manipulation} These techniques are typically designed for purposes such as visual effects, art creation, and augmented reality (AR). 
An approach closely aligned with our use case involves lighting estimation and virtual object insertion \cite{hold2017deep, hold2019deep, zhang2019all, li2020inverse, wang2021learning, wang2022neural}. Models in this category estimate scene lighting conditions and produce lighting representations, such as spherical functions~\cite{li2020inverse}, sky models~\cite{hold2019deep,hold2017deep,zhang2019all}, environment maps~\cite{gardner2017learning,legendre2019deeplight}, and volumetric representations \cite{srinivasan2020lighthouse,wang2021learning,wang2022neural}. With the estimated lighting information, virtual objects created by CG artists can be realistically inserted into images\cite{li2020inverse}, and lighting effects, such as cast shadows, can be properly handled~\cite{wang2022neural}. 
Our work distinguishes itself by addressing lighting estimation for real-world driving scenarios in an end-to-end manner and facilitating virtual object insertion across multiple cameras with realistic object placement, and ground truth generation. 
\begin{figure*}
\centering
\includegraphics[width=\linewidth]{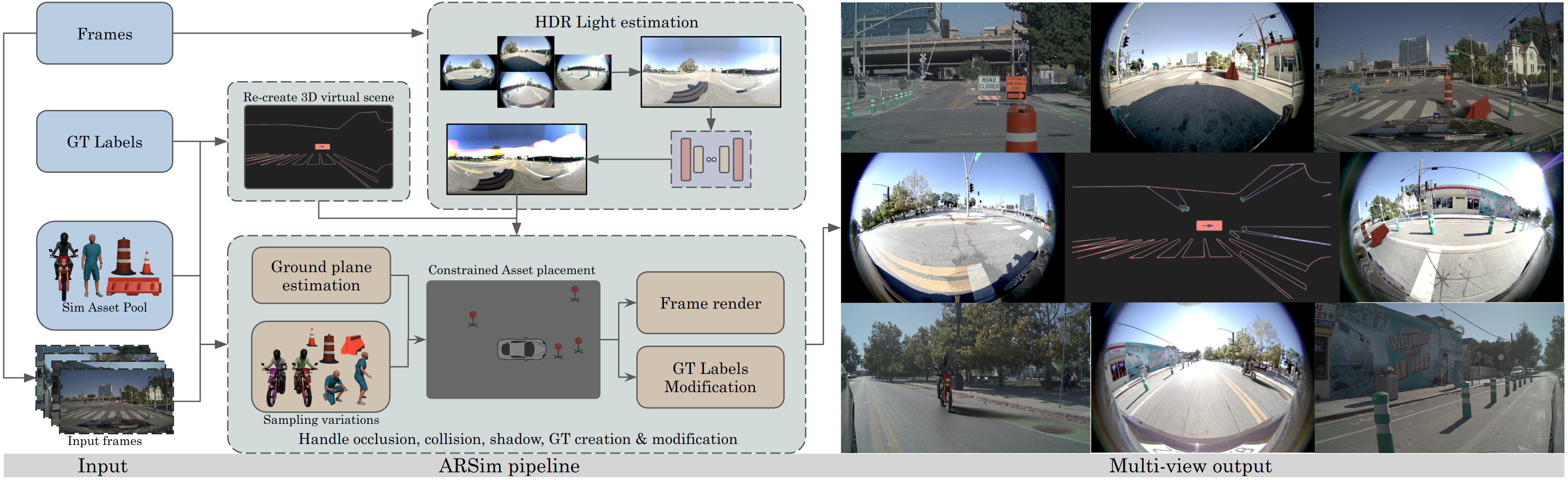}
\caption{(Left) An overview of ARSim's high-level block diagram. (Right) Example multi-view consistent data generated by ARSim with groundtruth generation and modification.}
\label{fig:arsim-block-diag}
\end{figure*}
\section{Problem Formulation}
\label{sec:problem-formulation}


In this paper, we target simulation-based synthetic data generation for the target application of autonomous vehicles. In the conventional simulation-based synthetic data generation \cite{richter2016playing, ros2016synthia, GuoCZZMWC20}, a gaming engine is used to create content and scenarios mimicking the use case under consideration. Such completely simulated scenarios, although offer more control in terms of variations, which is essential in generating a larger distribution of the data, come with a larger distribution gap due to a limited quantity of 3D environments and lacking photo-realistic quality. We aim to reduce these content and sensor domain gaps by using real data as background and only simulating what is needed. Instead of generating the entire scene (foreground, background), we render only the synthetic asset in each of the camera frames on top of real drive data.

\subsubsection*{Problem Setup}
During the training of an AV perception task, we are provided with $n_i$ number of samples, denoted as $S^i=\left\{\left(\mathbf{x}_k, y_k\right)\right\}_{k=1}^{n_i} \sim\left(\mathcal{D}_{i}\right)^{n_i}$, drawn from the distribution $\mathcal{D}_{i}$. Subsequently, a classifier trained on $S^i$ undergoes testing on $n_j$ samples, represented by $S^j=\left\{\left(\mathbf{x}_k, y_k\right)\right\}_{k=1}^{n_j} \sim\left(\mathcal{D}_{j}\right)^{n_j}$, sampled from $\mathcal{D}{j}$. Both the training and testing datasets can be viewed as subsets of the larger dataset sampled from the unbalanced distribution $\mathcal{D}$. Given that this distribution mirrors real-world conditions, it exhibits a long-tail distribution for less frequent occurrences. Consequently, due to this inherent imbalance, some samples or variations thereof may not exist in the training data, potentially leading to a covariate shift between the two subdomains, i.e., $P_{\mathcal{D}_i}(\mathbf{x}) \neq P_{\mathcal{D}_j}(\mathbf{x})$.

\subsubsection*{Domain Randomization}
One way to mitigate the long tail distribution is to sample from $\mathcal{D}_{i}$ in such a way that we oversample the less frequent cases. But this won't make the distribution any more balanced than it already is. The other approach is to make use of samples from a simulated data distribution $\mathcal{D}_{sim}$ which is intentionally generated in such a way that when merged with the samples from real data sampled from $\mathcal{D}_i$, will yield a more balanced distribution. This distribution however will introduce another covariate shift with $\mathcal{D}_{j}$, due to reduced photo realism in $\mathcal{D}_{sim}$.
To mitigate the covariate shift between $\mathcal{D}_{i}$ and $\mathcal{D}_{j}$, we propose a hybrid solution combining domain adaptation and domain randomization. Drawing from the Structural Causal Model (SCM) framework \cite{ilse2021selecting}, the data generation process for simulating images $\mathbf{x}$ with ground-truth labels $y$ can be expressed as $\mathbf{x} := f_{X}(h_{\mathcal{D}_{sim}}, h_{y})$, where the generated images are a function of domain attribute $h_{\mathcal{D}_{sim}}$ and label attribute $h_{y}$. Domain attributes encompass dataset characteristics devoid of direct causal relationships with the predicted outcome, including background landscape, brightness/contrast, sharpness, and perspective. Conversely, label attributes directly influence the predicted outcome and encompass features such as object shape and texture within an image.

\subsubsection*{Proposed Approach}
For many machine learning models, domain invariance is often overlooked, resulting in a potential spurious correlation between the domain attributes $h_{\mathcal{D}}$ and the label $y$. Domain randomization serves to mitigate this correlation by generating a wide variation of domain attributes in the simulated data. In our proposed approach, we combine domain adaptation and randomization, where a portion of the domain attributes (non-object attributes, e.g., background landscapes, lighting conditions, object shadow) is inferred from the real distribution $\mathcal{D}_{i}$, while simulation-based domain randomization is applied to other domain attributes (object attributes, e.g., shape, texture, color variations of the object). In essence, instead of simulating the entire spectrum, we focus on simulating what is necessary for domain randomization, thereby impacting only a small segment of the distribution. 
For the rest of the paper, we denote the data generated through simulation-based domain randomization as VRSim (Virtual Reality-based Simulated data) and the data generated through the proposed approach as ARSim (Augmented Reality-based Simulated data).

\section{ARSim Data Augmentation Pipeline}
\label{sec:method}

In this section, we present the modular ARSim pipeline highlighting its key components. The input to the pipeline is a real drive dataset, which is then processed for recreating a minimilast 3D virtual scene and inferring light distribution. Based on the long-tail distribution to be addressed, 3D assets are placed in the scene following the real-life constraints (collision, occlusion, ground placement etc). The objects are rendered across the different cameras and composited with the real images. Finally the pipeline is completed with generating the associating ground truth labels. 
The complete pipeline can be seen in Fig. \ref{fig:arsim-block-diag}.

\subsubsection*{Input Data Selection}
The first step for creating realistic augmentation data is to select the right input data. The realistic augmentation pipeline does not change the data format, outputting the data in the same format as that of the input data. A dataset can be used as input for the realistic augmentation if it has input frames with significant coverage around the ego car (essential for accurate light estimation), and the availability of intrinsic and extrinsic parameters of the camera used to generate frames. These two annotation pieces are essential for physically viable and photo-realistic augmentation to work. Other annotation information can be useful and increase the quality of the augmentation.

\subsubsection*{Scene Recreation}
For proper scene-aware placement of the 3D synthetic assets, additional annotations from the input data can be used. The more information we have from the input scene the richer this virtual 3D environment. The virtual scene can be as simple as 3D cuboids around the ego car and as complex as a NeRF scene (which requires significantly more amount of data compared to the former). To show the effectiveness of the realistic augmentation pipeline, we consider simple annotated datasets with the 3D cuboid and drivable freespace ground truth. The extrinsic and intrinsic parameters of the camera sensors mounted on the ego car during data collection are used to create camera objects within this virtual scene, which will be used to render the frame from each camera. We support both fisheye and non-fisheye camera models for broader coverage. Once this 3D virtual scene is set, the next important thing is to assign it a light distribution.

\subsubsection*{Lighting Estimation} 
As part of the re-created 3D virtual scene, our method estimates an environment map from a set of images captured by multiple cameras positioned around the ego vehicle. 
The input images often provide significant coverage of the surrounding views, but often have a limited dynamic range and are unable to cover the High Dynamic Range (HDR) intensity profile of outdoor scenes adequately, which are essential for generating lighting effects such as shadows. 
To address this, we make use of an encoder-decoder neural network trained with a dataset of HDR panoramas. The network is designed to transform an input LDR panorama to its corresponding HDR image. 
Specifically, the encoder maps the input image into three feature vectors following \cite{wang2022neural}, including the peak direction $\mathbf{f}_\text{d} \in\mathbb{R}^3$, peak intensity $\mathbf{f}_\text{i} \in\mathbb{R}^3$, and a latent vector $\mathbf{f}_\text{latent} \in\mathbb{R}^{64}$. These feature vectors are then processed by the decoder to output the HDR panorama. 
During inference time, the input images are unprojected onto an equi-rectangular panorama based on calibrated camera intrinsic and extrinsic parameters and then fused and fed into the encoder to predict the HDR environment map. We refer to the Supplement for more details.

\subsubsection*{Asset Placement} 
The placement of the synthetic assets should be viable and realistic, as seen by the camera objects. Hence considerations such as allowable region, object collision, and object occlusion must be made. 
Based on the specific use case, we can augment the assets based on the region of interest defined by the user if the scene structure allows it (such as objects along the ego lanes, far-away objects, and more). Based on the region of interest, a 3D location is randomly selected for each instance, and an attempt is made to place the synthetic assets at that position in the 3D virtual scene. The synthetic asset placed is checked for collision with existing objects in the scene. The asset is also checked for occlusion as viewed from each camera object since it depends not only on the asset's 3D location but also on the individual viewpoints. 
Based on the scene information available from the input data, we render the 3D synthetic asset for partial occlusion (say if the depth map or segmentation map is available). 

\subsubsection*{Scene Rendering}
Now that the virtual 3D environment is set, we render the frames from camera object only rendering the synthetically placed assets. A virtual plane is used as a shadow catcher for these assets, and for each camera frame, the object and the shadow are rendered separately. This separation between the object and the shadow map offers more control for realism. Post-processing is applied to these maps to adjust the shadow strength (shadow map), simulate basic sensor models (object map), and create a wide variation in the saturation of these rendered objects (object map). The post-processed object and shadow map are then alpha-composited with the real frame to generate the final augmented output.

\subsubsection*{Ground Truth Generation and Modification}
The concluding phase of the pipeline encompasses the creation of ground truth data corresponding to the augmented objects. This task entails the generation of both 3D and 2D bounding boxes along with the calculation of pertinent attributes associated with these bounding boxes. Additionally, it is imperative to update the existing input ground truth labels in most scenarios due to the augmentation of these assets. For instance, the addition of a synthetic object may affect the visibility of a real object cuboid. Furthermore, adjustments may be necessary at the scene level, such as modifying free space in certain cases. Upon completion of the ground truth generation process and the requisite label modifications, the dataset becomes available for training perception networks.

\begin{figure*}[t]
  \centering
  \begin{subfigure}[b]{0.64\linewidth}
  \includegraphics[width=\linewidth]{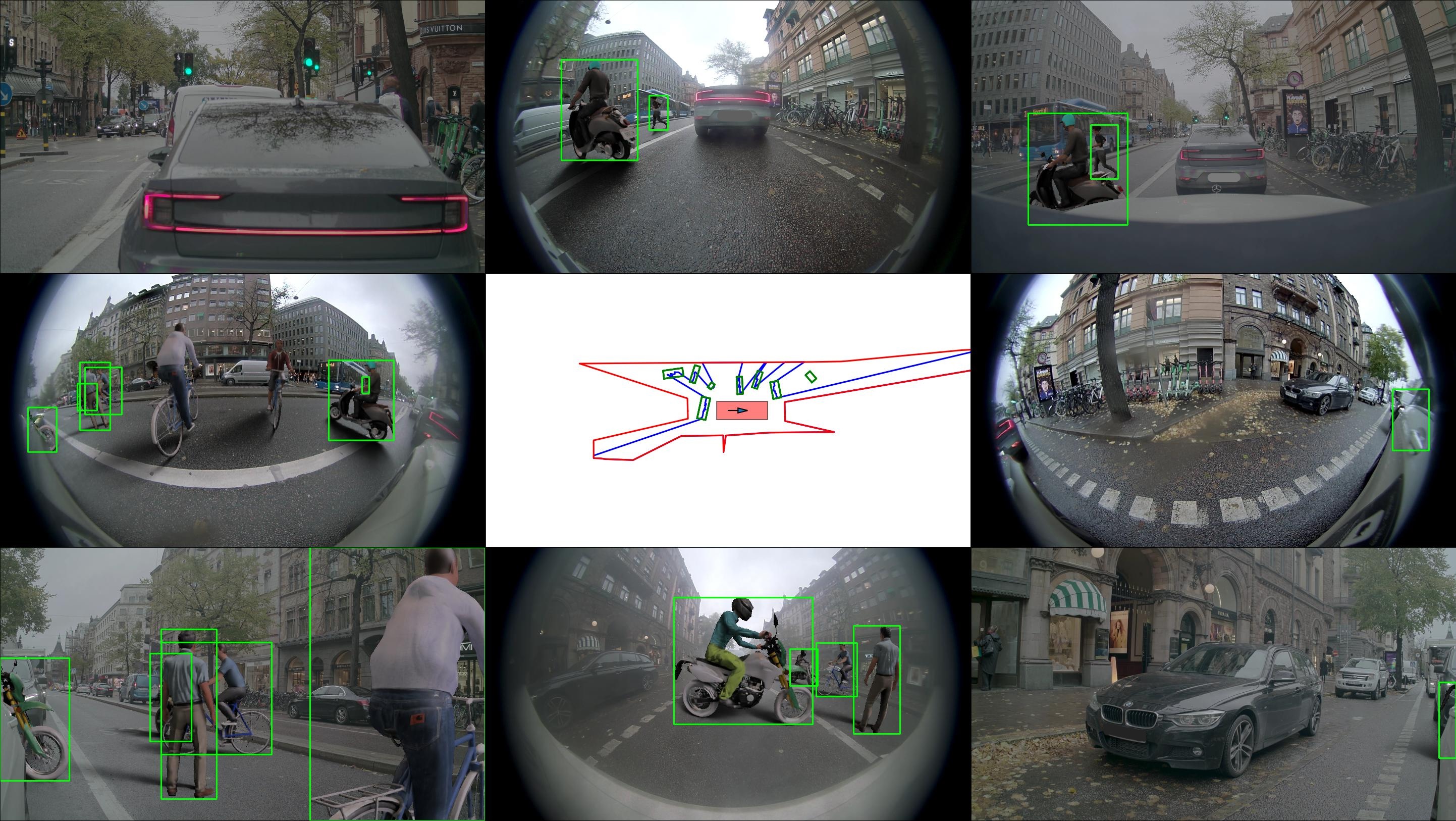}
\caption{Example obstacle ARSim data with multi-view consistency.}
\label{fig:obstacle-sample}
  \end{subfigure}
  \hfill
  \begin{subfigure}[b]{0.32\linewidth}
  \includegraphics[width=\linewidth]{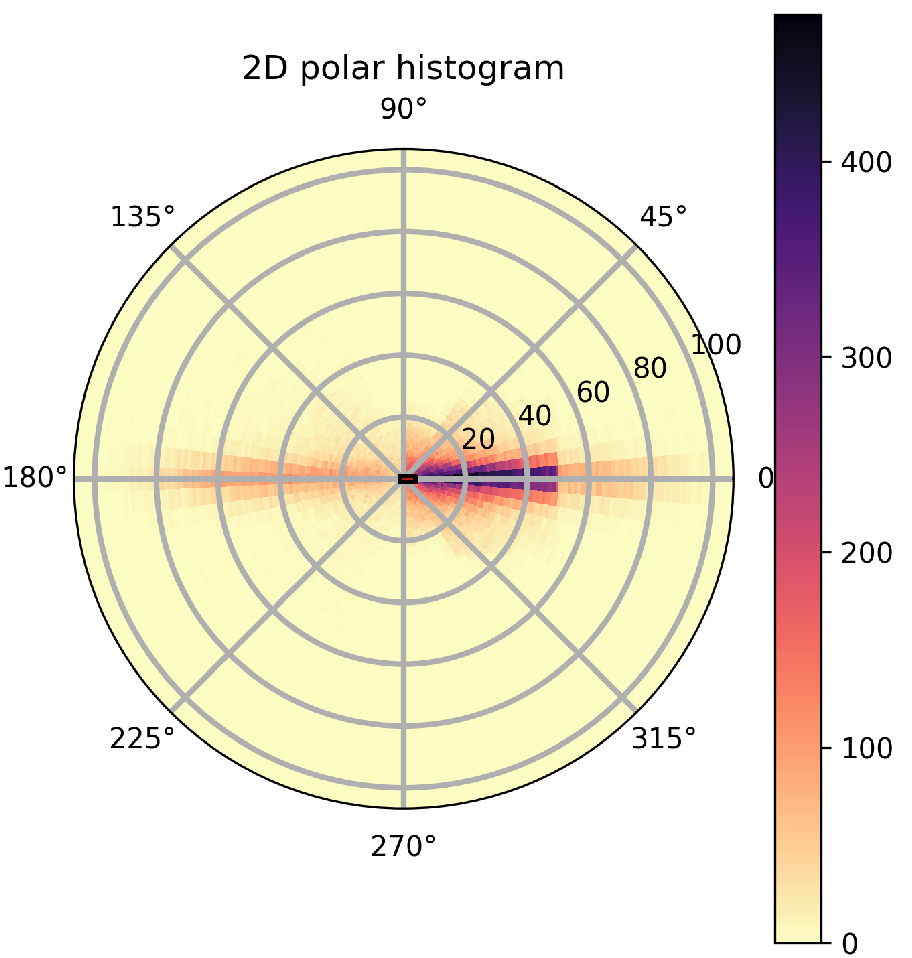}
    \caption{Spatial distribution of the assets augmented}
    \label{fig:obstacle-distribution}
  \end{subfigure}
  \caption{ARSim VRU data for improving obstacle detection}
  \label{fig:obstacle-data}
\end{figure*}

\begin{table}[tb]
  \centering
\begin{tabular}{lcccc}
\toprule
Dataset     & Scenes &  Person & Biker \\
\midrule
\texttt{Real(train)}           & 1M	                                & 	2.2M                & 	290K                           \\
\texttt{Real(test)}           & 306K	                                & 	266K                & 	29K                           \\
\texttt{ARSim}     & 140K                                & 186K                & 370K                           \\
\bottomrule   
\end{tabular}
  \caption{Statistics of the dataset used for obstacle detection improvement
  }
    \label{tab:obstacle-data}
\end{table}

\begin{table*}
\centering
\begin{tabular}{@{}lccccccccc@{}}
\toprule
{Class}& \multicolumn{2}{c}{Average Precision $\uparrow$} & \multicolumn{2}{c}{Fscore $\uparrow$} & \multicolumn{2}{c}{Position Err(m) $\downarrow$} & \multicolumn{2}{c}{Yaw Err($^{\circ}$) $\downarrow$} \\ \midrule
{} & {Real}       & {Real+ARSim}       & {Real}     & {Real+ARSim}      & Real   & Real+ARSim & Real   & Real+ARSim   \\ 
\cmidrule{2-9}
Biker             & 0.828	 & \textbf{0.842}            & 0.82    & \textbf{0.822}        & 0.769	 & \textbf{0.739}    & 9.0		&  \textbf{7.4}    \\
Person            & 0.807	 & \textbf{0.818}            & 0.794   & \textbf{0.797}          & 0.755 & \textbf{0.701}   & 26.9		&  \textbf{25.9}       \\ 
\bottomrule
\end{tabular}
\caption{Improvement in obstacle detection metrics for VRU classes. The arrows ($\uparrow, \downarrow$) indicate the desired direction of improvement for the metric}
\label{tab:obstacle-results}
\end{table*}

\begin{figure*}
  \centering
  \begin{subfigure}{0.52\linewidth}
  \includegraphics[width=\linewidth]{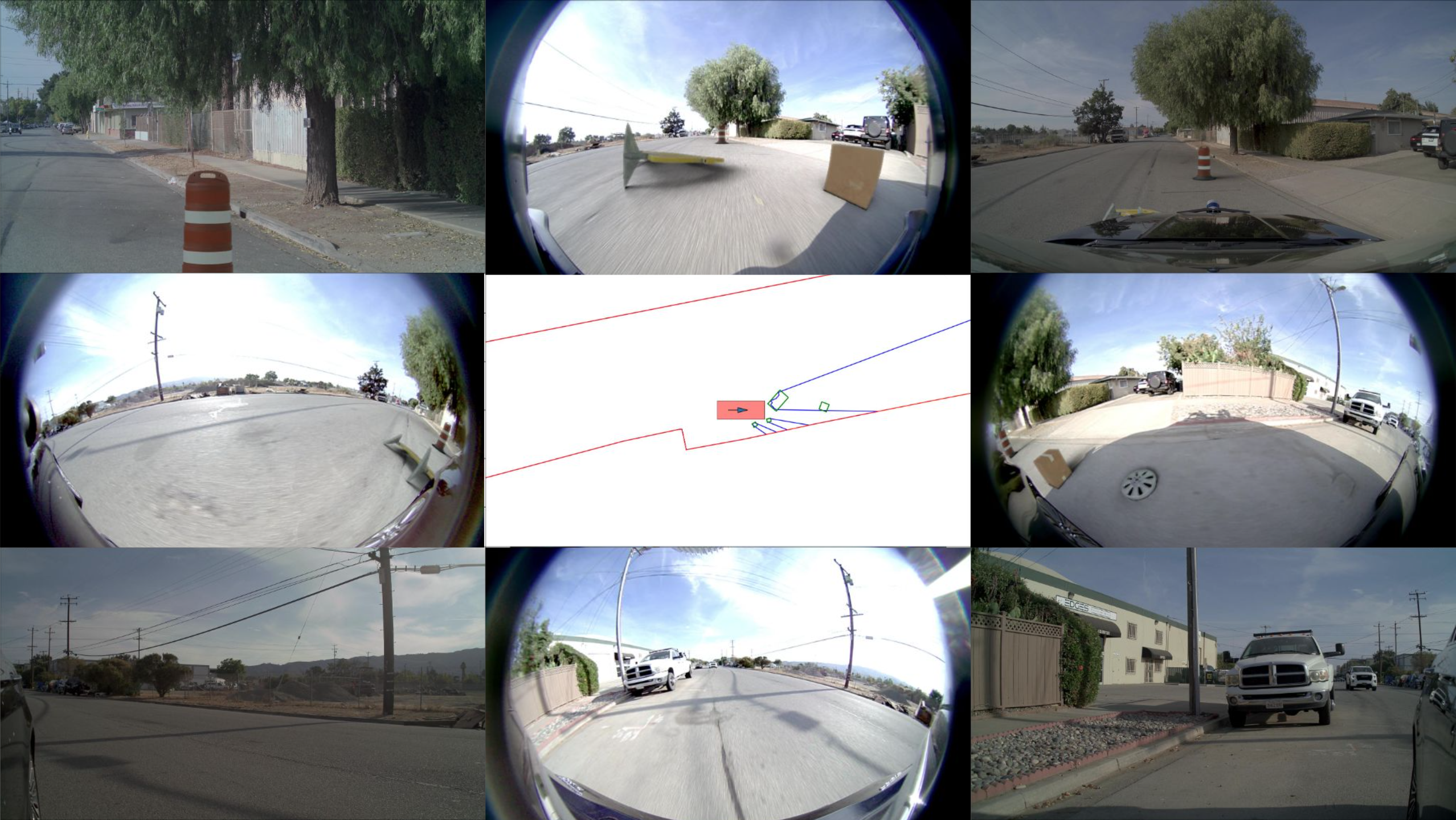}
\caption{Example freespace ARSim data with multi-view consistency.}
\label{fig:freespace-sample}
  \end{subfigure}
  \hfill
  \begin{subfigure}{0.45\linewidth}
  \includegraphics[width=\linewidth]{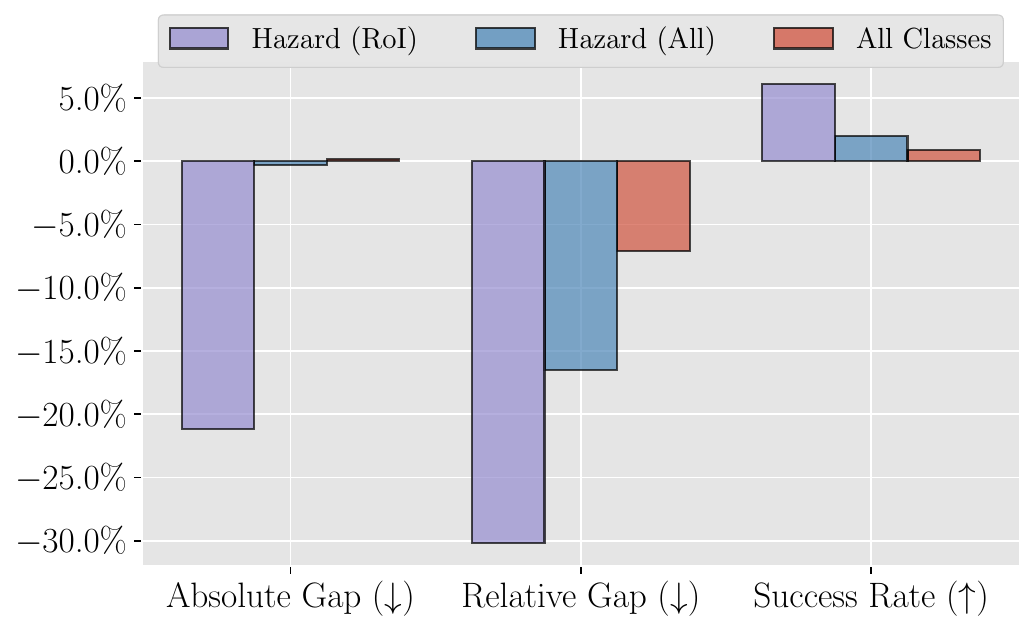}
    \caption{Improvement in freespace performance metrics, model trained on Real+ARSim vs Real}
    \label{fig:freespace-result-kpi}
  \end{subfigure}
  \caption{Impact of ARSim data on freespace detection}
  \label{fig:freespace-result}
\end{figure*}

\begin{table*}
\centering
\begin{tabular}{@{}ccccccc@{}}
\toprule
Dataset & Abs. Gap (m)$\downarrow$ & Rel. Gap (\%)$\downarrow$ & Success Rate (\%)$\uparrow$ & Precision$\uparrow$& Recall$\uparrow$\\ \midrule
\texttt{Real Only}                & 1.280                  & 29.62                & 69.49                & \textbf{0.745}             & 0.418           \\
\texttt{Real+ARSim}              & \textbf{1.009}                 & \textbf{20.68 }               & \textbf{73.73}               & 0.727              & \textbf{0.471}           \\ \bottomrule
\end{tabular}
\caption{Improvement in freespace performance metrics with ARSim dataset on real unseen test data for hazard class in a circular region of radius 10m around ego car.}
\label{tab:freespace}
\end{table*}

\section{Experiments}
\label{sec:results}


It's crucial to emphasize that the generation of the ARSim dataset is independent of the specific downstream training model employed. The choice of the training model is inconsequential and can be considered a black box. Instead of presenting absolute performance metrics, we opt to discuss relative improvement as a means to quantify the effectiveness of incorporating the ARSim dataset alongside real data. For the experiments outlined in this context, we utilize the NVAutonet model and training architecture described in \cite{pham2024nvautonet}. In the NVAutoNet framework, synchronized camera images are employed as input for predicting 3D signals, encompassing obstacles, freespaces, and parking. The initial step involves utilizing surround view images as input to CNN backbones, extracting essential 2D features. These features are then elevated and fused into a cohesive Bird's Eye View (BEV) feature map. The ensuing BEV features undergo further encoding through a dedicated BEV backbone. Finally, the system employs various 3D perception heads to make predictions for the 3D signals, completing the comprehensive process of 3D signal forecasting for objects like obstacles, freespaces, and parking.

For each of the target applications outlined below, we train NVAutoNet as a single-task problem.
Additionally, our base dataset consists of an in-house multi-camera dataset that includes essential sky coverage through fisheye cameras. Notably, other open-source datasets like nuScenes \cite{caesar2020nuscenes} or the Waymo open dataset \cite{sun2020scalability} do not incorporate fisheye cameras and/or lack a 360-degree field of view (FOV) coverage. Consequently, they lack the crucial sky coverage essential for accurate light estimation. We aim to enhance the perception networks of AVs by identifying specific areas that cater to a diverse range of applications and problem scenarios. To achieve this, we employ the strategic implementation of data augmentation using ARSim discussed below.


\subsection{Improving Obstacle Detection:}

Detecting 3D objects is crucial for AVs, encompassing vehicles and Vulnerable Road Users (VRUs) like pedestrians, cyclists, and motorcyclists. While there's an abundance of vehicle data from real drives, VRU data is scarce due to safety and cost challenges. To enhance VRU detection, we utilize the ARSim dataset, derived from an in-house real drive dataset featuring 1 million scenes. We randomly sample 140k scenes to generate ARSim data, incorporating over 40 synthetic 3D objects for bikers and pedestrians. These objects are strategically placed around the ego car, with variations in color and position to simulate real-world scenarios. Figure \ref{fig:obstacle-sample} showcases an example ARSim scene with synthetic VRUs, while Figure \ref{fig:obstacle-distribution} depicts asset spatial distribution. Relevant statistics are outlined in Table \ref{tab:obstacle-data}.

\begin{figure*}
  \centering
  \begin{subfigure}{0.48\linewidth}
  \includegraphics[width=\linewidth]{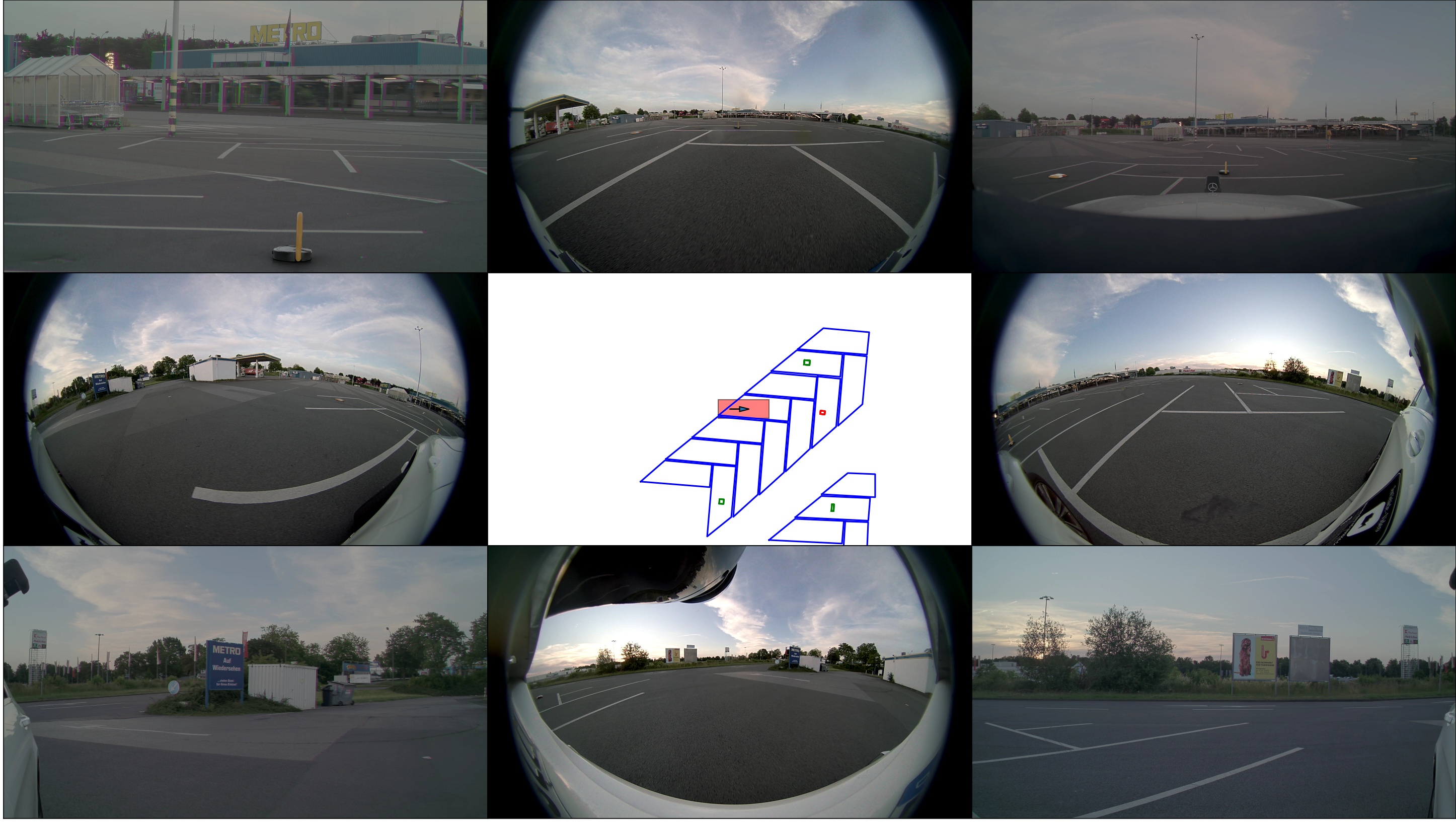}
    \caption{Example synthetic ground lock augmented on real scenes with multi-view consistency.}
    \label{fig:parking-sample}
  \end{subfigure}
  \hfill
  \begin{subfigure}{0.48\linewidth}
  \includegraphics[width=\linewidth]{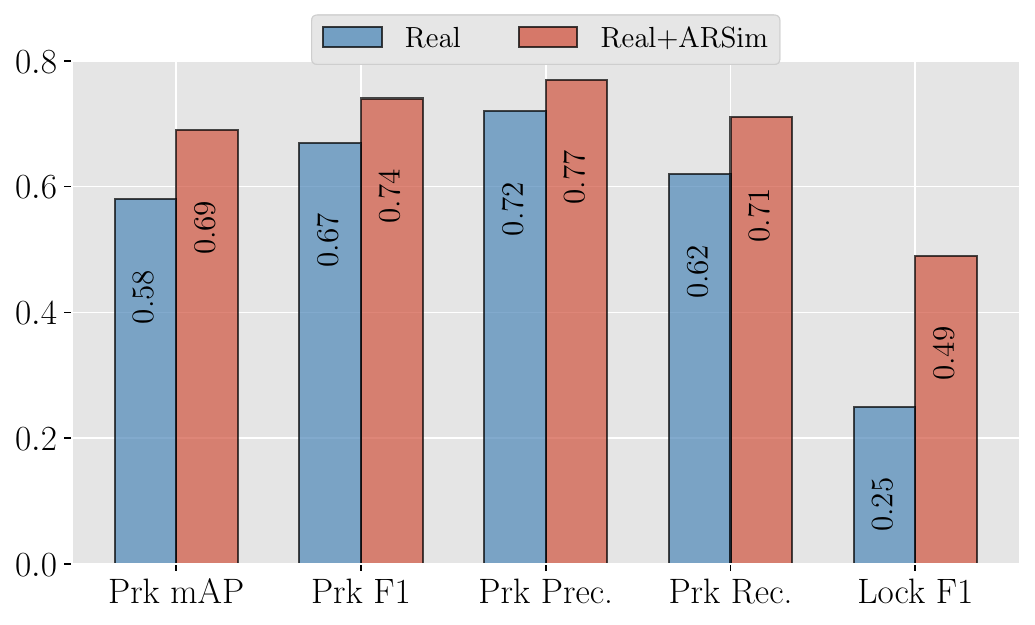}
    \caption{Improvement in parking performance metrics using ARSim}
    \label{fig:parking-result}
  \end{subfigure}
  \caption{Improvement in person/biker class performance metrics using ARSim}
  \label{fig:parking}
\end{figure*}

\begin{table*}
  \centering
\begin{tabular}{lccccc}
\toprule
Dataset    & Scenes & Total Objects & Locked GLs & Unlocked GLs & GLs/scene \\
\midrule
\texttt{Real (train)}                 & 967K               & 24.8K                        & 18.5K                     & 6.3K                         & 0.03                  \\
\texttt{ARSim}                & 52K               & 131K                      & 60.1K                     & 70.8K                        & 2.51                  \\
\texttt{Real+ARSim}           & 1.01M            & 155.8K                    & 78.7K                     & 77K                        & 0.15                  \\
\texttt{Real (test)}             & 161.5K               & 2K                         & 1.2K                      & 0.8K                          & 0.01                  \\
\bottomrule
\end{tabular}
  \caption{Statistics of the dataset used for ground lock detection in parking model
  }
    \label{tab:parking-data}
\end{table*}

Two NVAutonet models were trained for the obstacle3d task using Real and Real+ARSim data. The performance of the model was evaluated on real unseen test data with 306K scenes. Table \ref{tab:obstacle-results} reports the average precision, F-score, position error, and yaw angle error for the targeted class of person and biker. Position error is the Euclidean distance between the ground truth and the predicted 3D cuboid center in meters, and yaw error is the absolute difference in the yaw angle of the ground truth and the predicted 3D cuboid. A predicted cuboid3D is said to be a true positive if the relative radial distance between its center and ground truth is less than 10\% and the cuboid orientation Angle difference is within 2 degrees. Table \ref{tab:obstacle-results} reports the necessary quantitative metrics. It can be seen that making use of the ARSim data not only improved the detection metrics (average precision, F-score) but also reduced the position/orientation errors for the cuboids. 
It is important to note that the small improvements in classification metrics observed when utilizing the ARSim data can be attributed to the intricate and diverse nature of real-world human data, alongside the already high performance of the specific quantitative metrics being targeted.


\subsection{Improving Freespace Detection}
Freespace refers to the area within road boundaries devoid of obstacles, playing a pivotal role in ensuring the safe navigation of autonomous vehicles. 
Much like 3D obstacle data, freespace data also exhibits a long-tail distribution, with minimal representation from hazard objects. Hazard objects encompass both road debris (stray objects frequently found on the road) and intentionally placed traffic objects. 
Our focus is on addressing the challenge of close-range freespace detection for hazard objects.

We utilized an in-house real dataset comprising 1.35 million scenes with associated freespace labels. Scenes from this real-drive dataset were randomly sampled to generate the ARSim dataset. In this process, approximately 75 instances of synthetic objects were employed, encompassing a diverse array of hazards such as boxes, tires, chairs, trash cans, debris, shopping carts, scooters, traffic poles, cones, barrels, barriers, triangles, and more. 
Multiple Synthetic objects were placed within a rectangular region centered around the ego car, featuring a longitudinal variation of $\pm12m$ and a lateral variation of $\pm6m$. The ARSim dataset was generated, producing a total of 100K scenes and incorporating 226K synthetic hazard objects. Fig \ref{fig:freespace-sample} shows a sample freespace ARSim data with 4 synthetic objects (barrel, barrier, box, tire rim) in front and right cameras.

Two models were trained for the freespace head in NVAutoNet: one using only real data and the other using both real and ARSim data. The freespace region is represented as a Radial Distance Map (RDM), composed of equiangular bins around the car, each containing a radial distance value to the closest obstacle and a semantic label. Table \ref{tab:freespace} presents performance metrics for the two models, evaluated on a separate real test dataset of 100k scenes. The absolute gap (Abs. gap) and relative gap (Rel. Gap) denotes average radial distance errors, while the success rate indicates the percentage of successfully estimated angular bins, where success is defined by a relative gap of less than 10\%. The arrows beside the metric indicate the desired direction of improvement. The models were targeted for hazard detection within a circular region with a radius of 10m around the ego car. Results show a reduction of over 20\% and 30\% in absolute and relative gap respectively, with improved recall for freespace bin classification to hazard class, albeit a slight impact on precision. Fig \ref{fig:freespace-result-kpi} illustrates the improvement achieved by incorporating the ARSim dataset across different evaluation scenarios.

\subsection{Improving Parking Detection for ground locks}
Automatic parking involves perception tasks such as detecting the parking spots, their availability, and then finally maneuvering safely to the spot and completing the parking. In some countries, the availability of a parking spot is determined by parking ground locks (GL), which are physical devices installed in the parking spots. When engaged, these locks create a physical barrier, thereby rendering the parking spot unusable for vehicles. 
Real datasets do not contain enough parking ground locks for training, hence our objective is to use ARSim-generated data to improve the detection of various ground locks.


Utilizing an in-house real-drive parking dataset, we randomly sample $\sim50K$ scenes to augment ground locks in both locked and unlocked states using 6 different instances of 3D synthetic ground locks. The ground locks were placed in the middle of the parking spot (with a Gaussian distribution for placement noise) by estimating a ground plane through these parking spots. A sample from the parknet ARSim data with ground locks can be seen in Fig \ref{fig:parking-sample}.

We trained NVAutoNet's parking head with Real and Real+ARSim datasets and tested on real-drive test data. The details of these datasets can be seen in Table \ref{tab:parking-data}. Fig. \ref{fig:parking-result} plots the performance metrics with and without using the ARSim dataset when evaluated on test data.  
Prk mAP, Prk F1 Prk Prec. and Prk Rec, are the mean average precision, F1 score, precision, and recall for the parking spot detection. Lock F1 is the F-score of a binary variable that signifies whether there is a lock within the parking spot or not. It can be seen that the detection of ground lock (Lock F1) was increased by 96\%. Moreover, adding ground locks also improved the detection of the parking spots as it provided an additional signal for parking spot detection.

\section{Ablation Test}
\label{sec:ablation}
\begin{figure*}
  \centering
  \begin{subfigure}{0.31\linewidth}
  \includegraphics[width=\linewidth]{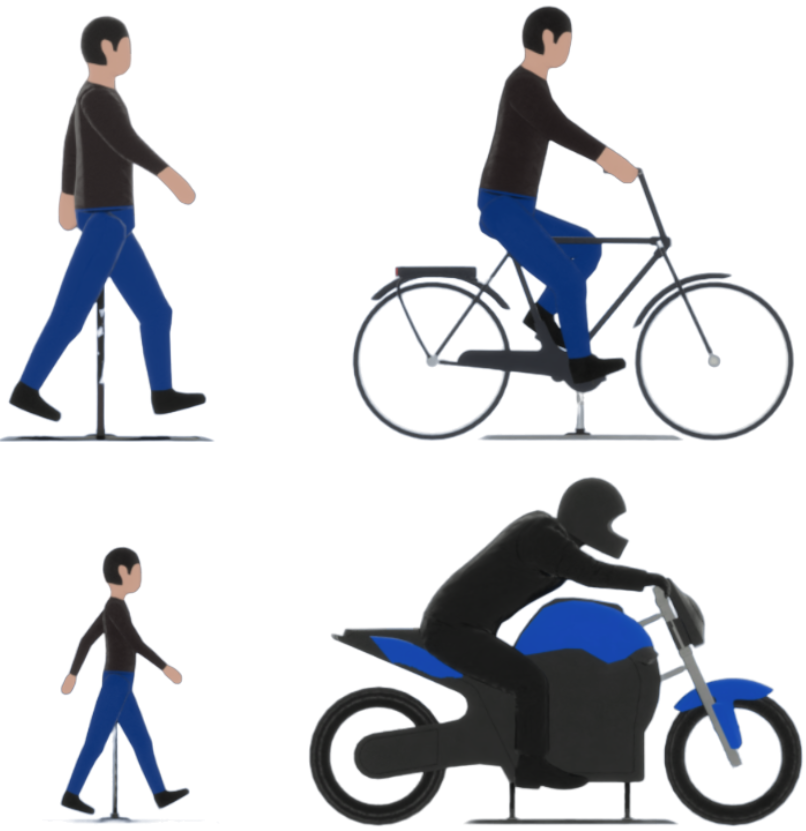}
    \caption{Euro NCAP dummies used for augmentation}
    \label{fig:ncap-dummies}
  \end{subfigure}
  \hfill
  \begin{subfigure}{0.65\linewidth}
\includegraphics[width=\linewidth]{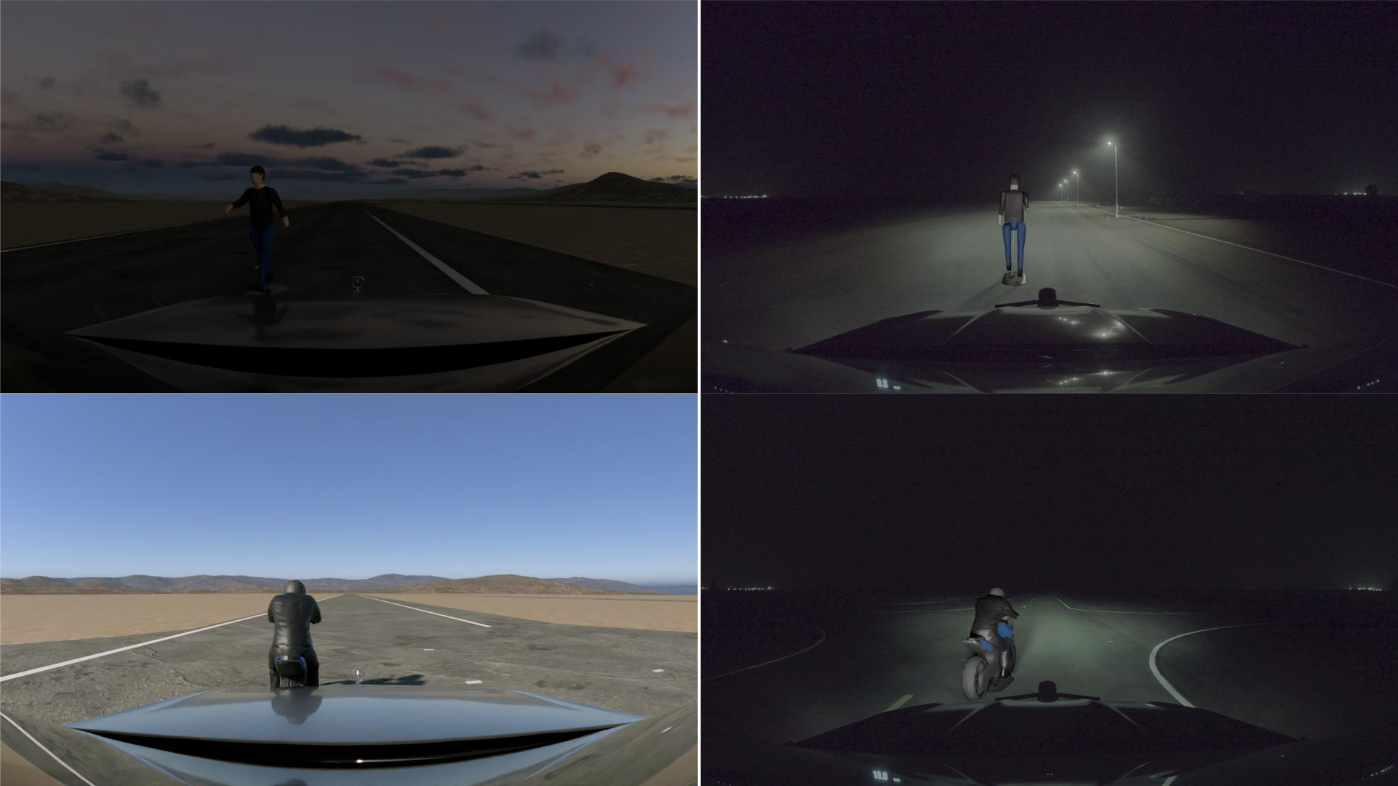}
\caption{Comparisons of the frames generated from VRSim (left) and ARSim (right)}
\label{fig:arsim-vrsim-demo}
  \end{subfigure}
  \caption{Sim data for NCAP ablation study}
  \label{fig:ablation}
\end{figure*}

In this section we compare the performance of models, one compensated with the VRSim dataset and the other with the proposed ARSim dataset. We consider the target application of Euro New Car Assessment Programs (NCAP) dummy detection. Euro NCAP is a set of safety protocols that determines the safety level of a new car.
The key perception task for this certification is 3D obstacle detection for pre-defined NCAP dummies as shown in Fig \ref{fig:ncap-dummies}.

An in-house real dataset with 2.2 million scenes is used as the base dataset. We generate 50K scenes of ARSim and VRSim datasets using four synthetic assets including child, adult, bicycle, and motorbike NCAP dummies shown in Fig \ref{fig:ncap-dummies} covering 21 scenarios defined in the Euro NCAP guideline. The camera frame from the generated ARSim and VRSim datasets can be seen in Fig. \ref{fig:arsim-vrsim-demo}.
Various NVAutoNet models for obstacle heads were trained, with different combinations of Real, VRSim, and ARSim datasets and were evaluated on a target dataset. The details of training and evaluation datasets can be seen in Tab. \ref{tab:ablation-data}

\begin{table}
  \centering
\begin{tabular}{lcccc}
\toprule
Dataset     & Scenes &  Person & Biker \\
\midrule
\texttt{Real (train:all)}           & 2.2M	                                & 	6.3M                & 	929K                           \\
\texttt{Real (target:NCAP)}     & 48K                                & 15K                & 3K                           \\
\texttt{ARSim50k}             & 51K                            & 31K                & 17K                           \\
\texttt{VRSim50k}             & 51K                              & 31K                & 17K                           \\
\bottomrule   
\end{tabular}
  \caption{Statistics of the dataset used for ARSim/VRSim ablation test
  }
  \label{tab:ablation-data}
\end{table}

\begin{table}[tb]

  \centering
\begin{tabular}{@{}ccccc@{}}
\toprule
\multicolumn{1}{l}{Class} & \multicolumn{2}{c}{AP} & \multicolumn{2}{c}{Fscore} \\ \midrule
                & VRSim  & ARSim & VRSim    & ARSim   \\
                \cmidrule{2-5}
Biker               & 0.02            & \textbf{0.322}          & 0.146            & \textbf{0.441}            \\

person                        & 0.102         & \textbf{0.52}           & 0.167             &  \textbf{0.552}            \\ 
\bottomrule
\end{tabular}
  \caption{Metrics evaluation of models on target dataset trained with sim-only datasets}
\label{tab:kpi-sim-only}
\end{table}

\subsubsection*{Train with Sim only:}
Tab. \ref{tab:kpi-sim-only} shows the average precision and Fscore for the classes Biker and Person for two models trained on VRSim and ARSim data only and evaluated on the real NCAP target dataset. It can be seen that the model trained on ARSim significantly outperforms the model trained with VRSim data. One of the reasons for this significant difference in performance is the reduced domain gap between ARSim and the real target data as compared to VRSim. It can be seen that even though the performance of the ARSim-trained model is relatively higher, it still is not good enough to be used independently. Hence a combination of real and sim datasets is always helpful in bridging this gap.
\subsubsection*{Train with Real+Sim:}
\begin{figure*}
  \centering
  \begin{subfigure}{0.49\linewidth}
  \includegraphics[width=\linewidth]{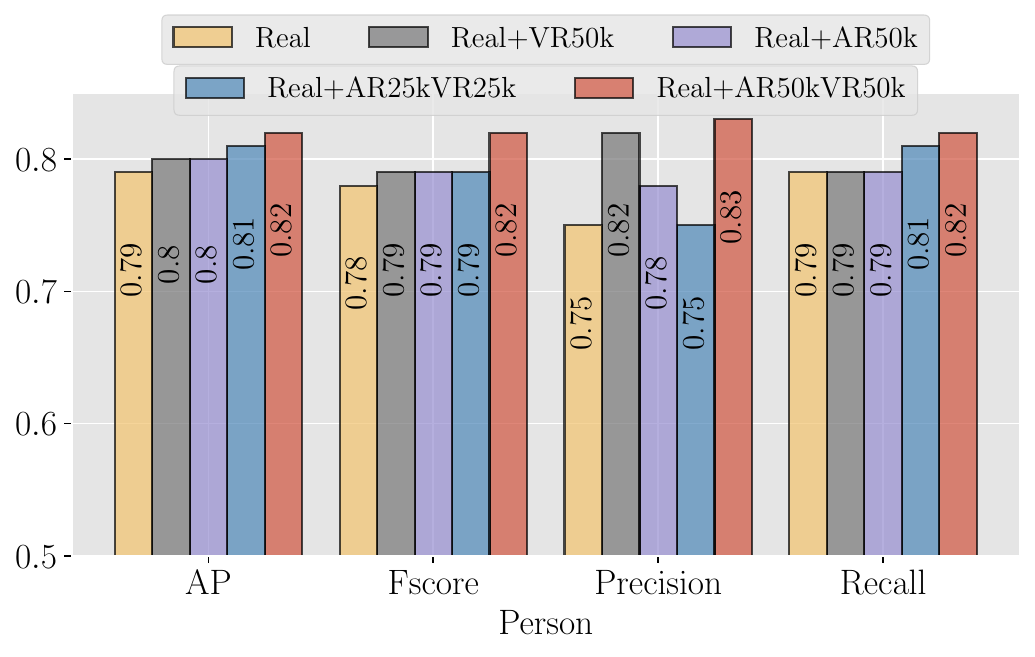}
    \label{fig:ablation-person}
  \end{subfigure}
  \hfill
  \begin{subfigure}{0.49\linewidth}
  \includegraphics[width=\linewidth]{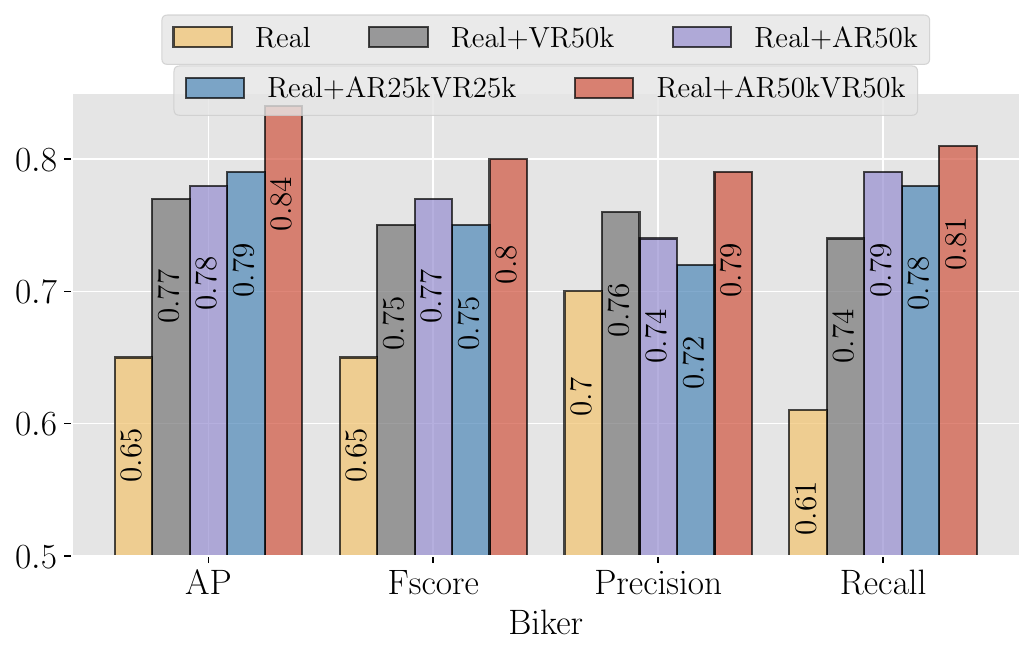}
    \label{fig:ablation-bike}
  \end{subfigure}
  \caption{Metric evaluation on target dataset with models trained on various combinations of real and sim data}
  \label{fig:ablation-result}
\end{figure*}

We train various models with multiple combinations of real and sim data and evaluate the performance on the target dataset. Fig. \ref{fig:ablation-result} plots the AP, Fscore, Precision, and Recall for the targeted classes of Biker and Person. Real+AR\textit{x}kVR\textit{y}k is a model trained with a combination of real, \textit{x} thousand scenes of ARSim and \textit{y} thousand scenes of VRSim data. It can be seen that introducing a small fraction of sim data improved the AP and F-score for both classes. ARSim tends to perform slightly better in AP, F-score, and recall, while VRSim is better in the precision metric. Using a combination of both the ARSim and VRSim data seems to be the best fit to improve the performance metrics. Moreover, introducing more sim data (Real+AR50k+VR50k) improves the metrics even further, suggesting room for improvement.

\section{Conclusion}

In this paper, we presented a novel augmented reality-based method designed to mitigate the covariate shift between real and simulated data within autonomous vehicle (AV) perception networks. This approach integrates domain adaptation and randomization strategies to effectively address the disparities between the two datasets. Specifically, we infer crucial domain attributes from real data, while leveraging simulation-based randomization techniques for other attributes. The proposed method estimates the lighting distribution from the input images and augments the 3D synthetic objects of interest with multi-view consistency to address the long tail distribution of the real dataset. Through our experiments, spanning obstacle, freespace, and parking training models, we have demonstrated significant performance improvements by incorporating both real and ARSim datasets. Notably, our method outperforms its VRSim counterpart, with even more remarkable enhancements observed when combining both datasets for training. This underscores the synergy between the two approaches and emphasizes the effectiveness of our augmented reality-based methodology. Importantly, our method is not confined to AV applications but can be adapted to various scenarios requiring data augmentation across multiple camera

\bibliographystyle{plain}
\bibliography{egbib}

\clearpage

\title{Supplementary Material}
\author{}
\maketitle

\begin{figure*}
\centering
\includegraphics[width=\linewidth]{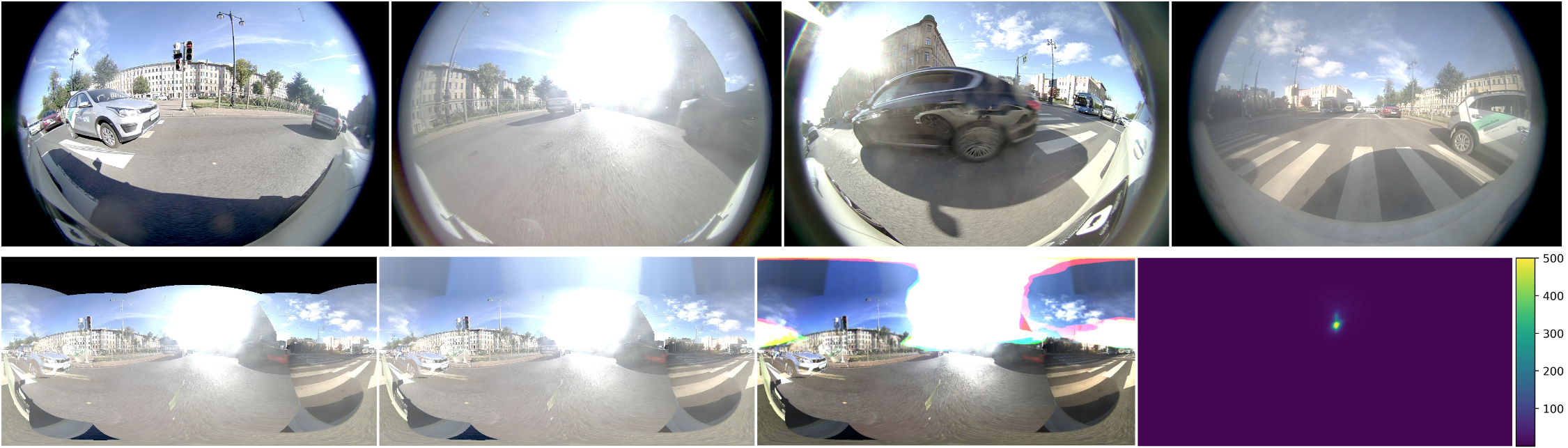}
\caption{Light estimation - left, front, right and rear fisheye cameras (top row), stitched panorama, inpainted panorama, HDR panorama, luminance of HDR panorama (bottom row) }
\label{fig:light_calc}
\end{figure*}

\appendix
\setcounter{section}{0}

This document supplements the main paper and provides additional implementation details (Sec.~\ref{sec:supp_implementation}), examples of generated ARSim data (Sec.~\ref{sec:supp_arsim_data}), and details on improving detection metrics using ARSim data for pre-trained networks (No new models were trained) (Sec.~\ref{sec:supp_arsim_improvement}).

\section{Implementation Details}
\label{sec:supp_implementation}

\subsection{Lighting estimation} 
The encoder-decoder architecture of our lighting estimation module follows the sky modeling network in~\cite{wang2022neural}. 
The encoder architecture is a ResNet50~\cite{resnet}. The input to the ResNet encoder is a 6-channel image, including an LDR panorama $\mathbf{I}_\text{LDR} \in \mathbb{R}^{H\times W \times3}$ and a positional encoding $\mathbf{I}_\text{pe} \in \mathbb{R}^{H\times W \times3}$ where each pixel contains its direction in equirectangular projection. 
The encoder outputs a set of sky feature vectors including the peak direction $\mathbf{f}_\text{d} \in \mathbb{R}^{3}$, peak intensity $\mathbf{f}_\text{i} \in \mathbb{R}^{3}$, and a latent vector $\mathbf{f}_\text{latent} \in \mathbb{R}^{64}$. 

The decoder is a 2D UNet~\cite{unet} that consumes the sky feature vectors. 
Before feeding into the decoder, the predicted peak direction $\mathbf{f}_\text{d}$ is converted as a 1-channel panorama $\mathbb{R}^{H\times W \times 1}$ encoding the peak direction 
$\mathbf{I}_\text{peak} = e^{100\,(\mathbf{I}_\text{pe}\cdot \mathbf{f}_\text{d} - 1)}$. 
The peak intensity $\mathbf{f}_\text{i}$ is converted as a 3-channel panorama $\mathbf{I}_\text{i} \in \mathbb{R}^{H\times W \times 3}$:
\begin{equation}
    \mathbf{I}_\text{i} = \left\{
    \begin{aligned}
        &\mathbf{f}_\text{i},      & \ \ \text{if} \ \mathbf{I}_\text{peak} \ge 0.98 \\
        &0,     & \text{Otherwise}
    \end{aligned}
    \right. 
    \label{eq:PeakIntensityEncoding}
\end{equation}
The decoder take as input a 7-channel panorama by concatenating $\mathbf{I}_\text{peak}, \mathbf{I}_\text{i}$ and $\mathbf{I}_\text{pe}$ and concatenate latent vector $\mathbf{f}_\text{latent}$ in the UNet latent space. The output of the decoder is an HDR panorama $\mathbf{I}_\text{HDR} \in \mathbb{R}^{H\times W \times 3}$. Following~\cite{wang2022neural}, we train the network on the HoliCity dataset and a set of 724 outdoor HDR panoramas. The panoramas are randomly rotated along the azimuth as a data augmentation.

During inference, we unproject the four input fisheye images onto panoramas with standard equi-rectangular projection and merge them into one panorama with maxpooling. The input fisheye images typically cover all or most of the panoramas, and we inpaint the panoramas~\cite{telea2004image} if there are empty pixels. The resulting panorama is then fed into the encoder-decoder architecture to predict the HDR panorama $\mathbf{I}_\text{HDR}$. The final environment map $\mathbf{I}_\text{envmap}$ fills saturated pixels with predicted HDR values and keeps the high-frequency details from the LDR panorama: 
\begin{equation}
    \mathbf{I}_\text{envmap} = \left\{
    \begin{aligned}
        &\mathbf{I}_\text{HDR},     & \ \ \text{if} \ \mathbf{I}_\text{LDR} \ge 0.9 \\
        &\mathbf{I}_\text{LDR},     & \text{Otherwise}
    \end{aligned}
    \right. 
\end{equation}

Fig. \ref{fig:light_calc} depicts the light estimation through an example.

\subsection{Ego light modeling}
\begin{figure*}
\centering
\includegraphics[width=1.0\linewidth]{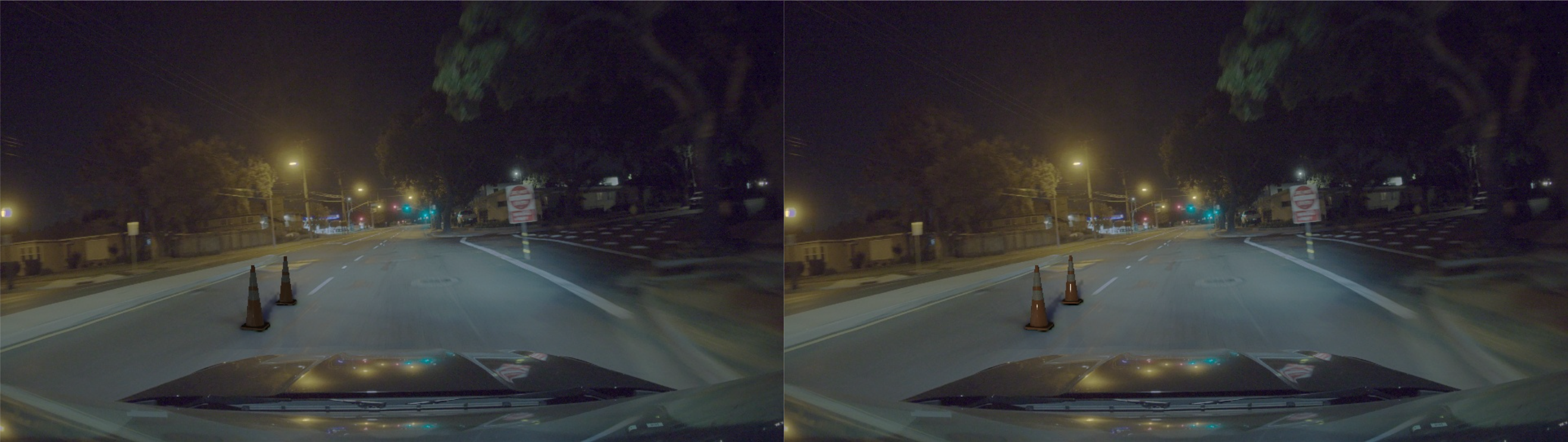}
\caption{Modelling the ego car headlights. (left) ARSim generated image without modeling the ego car headlight. (right) ARSim generated image after modeling the ego car headlights.}
\label{fig:headlight-activation}
\end{figure*}

In addition to the 360-degree HDR panorama mentioned earlier for illuminating the scene, we incorporate other light sources specifically for night-time scenes. The ego car's headlights and rear lights are simulated as conic point light sources with corresponding colors. These lights are triggered when the luminance of the HDR map ($ 0.2126 \times R + 0.7152 \times G + 0.0722 \times B$) \cite{enwiki:luminance} falls below a predefined threshold of 0.5. Figure \ref{fig:headlight-activation} displays the ARSim data captured by the front-facing camera both before and after the activation of the ego headlights. Incorporating the modeling of the ego lights enhances the realism of rendered assets in night-time scenes.


\begin{figure*}[hbtp]
\centering
\includegraphics[width=1.0\linewidth]{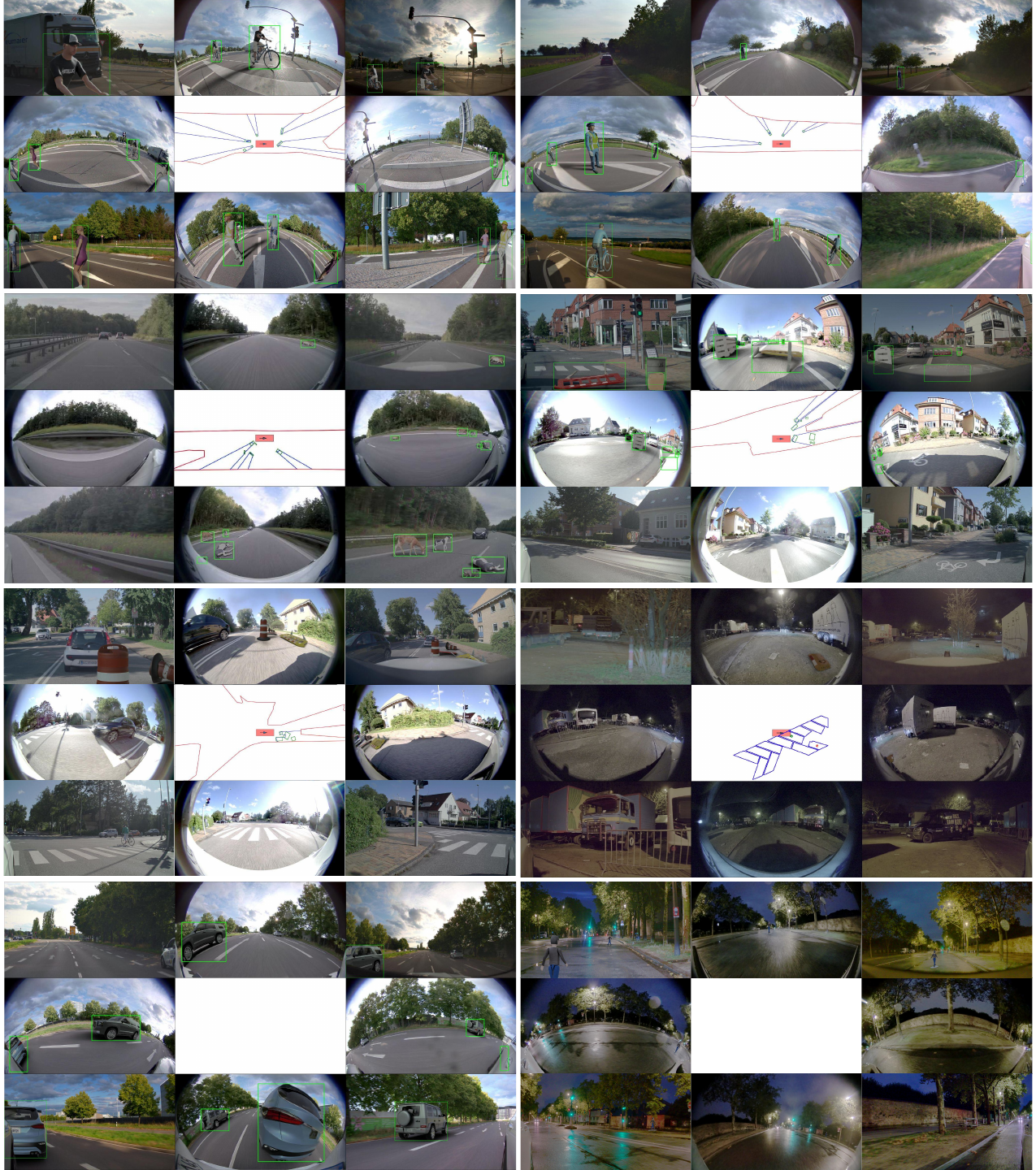}
\caption{Examples of ARSim multi-view consistent data}
\label{fig:examples}
\end{figure*}

\section{Generated ARSim data}
\label{sec:supp_arsim_data}
\subsection{Asset placement}

We employ stratified sampling to position 3D synthetic assets within the scene for rendering. To ensure balanced representation across target DNNs, the synthetic 3D assets are organized into categories. Given that some categories may contain more instances than others, uniform sampling could result in a skewed distribution. Thus, we establish a categorical distribution $p(n)$ representing the number of synthetic assets $n \in\{1, 2, \ldots, N\}$ to be placed in a scene (e.g., placing [1, 2, 3, 4] assets with probabilities [0.4, 0.3, 0.2, 0.2]), irrespective of available asset categories. Subsequently, we define another categorical distribution for asset categories $p(g)$, where $g$ denotes the asset group (e.g., group of pedestrians), $g \in\{1, 2,  \ldots, G\}$, and $G$ represents the total number of asset groups ($\sum_{g=1}^{G}p(g) = 1$). Prior to placing an asset, we sample the number of assets $n$ from $p(n)$. This $n$ is then allocated across group $g$ based on $p(g)$ and rounded to the nearest integer to yield $n_g$ (i.e., the number of assets from asset group $g$ to be placed). Finally, we uniformly sample $n_g$ instances from group $g$ to position them within the scene. This approach ensures an average of $n_{avg} = \sum_{n=1}^{N}p(n)*n$ assets per scene, with each asset group $g$ having $n_{avg}*p(g)$ occurrences per scene.
\subsection{Range of scenarios}
It can be seen in Figure \ref{fig:examples} that our proposed approach of ARSim, aimed at enhancing real data by augmenting 3D synthetic assets with multi-view consistency, encompasses a broad spectrum of scenarios. This coverage extends across different types of cameras, a vast asset pool, accurate lighting simulations, and diverse time settings. Regarding camera types, our system supports pinhole, standard, and fisheye lens cameras using f-theta lens model \cite{enwiki:fisheye}, with the latter adept at rendering 3D synthetic objects with the necessary distortion along the outer edges, ensuring realism (as evidenced by the example of the synthetic car in the rear fisheye camera shown in the last row and first column). Our approach caters to a comprehensive range of 3D synthetic assets, including VRUs, animals, cars, hazard objects, groundlocks, and New Car Assessment Programs (NCAP) dummies. Importantly, the generated ARSim data showcases realistic lighting and shadows, seamlessly integrating with other elements within the scene. These examples also illustrate various daytime conditions, spanning from cloudy and sunny weather to nighttime and rainy environments, further highlighting the versatility and applicability of our approach across diverse scenarios.

\begin{figure*}
\centering
\includegraphics[width=0.6\linewidth]{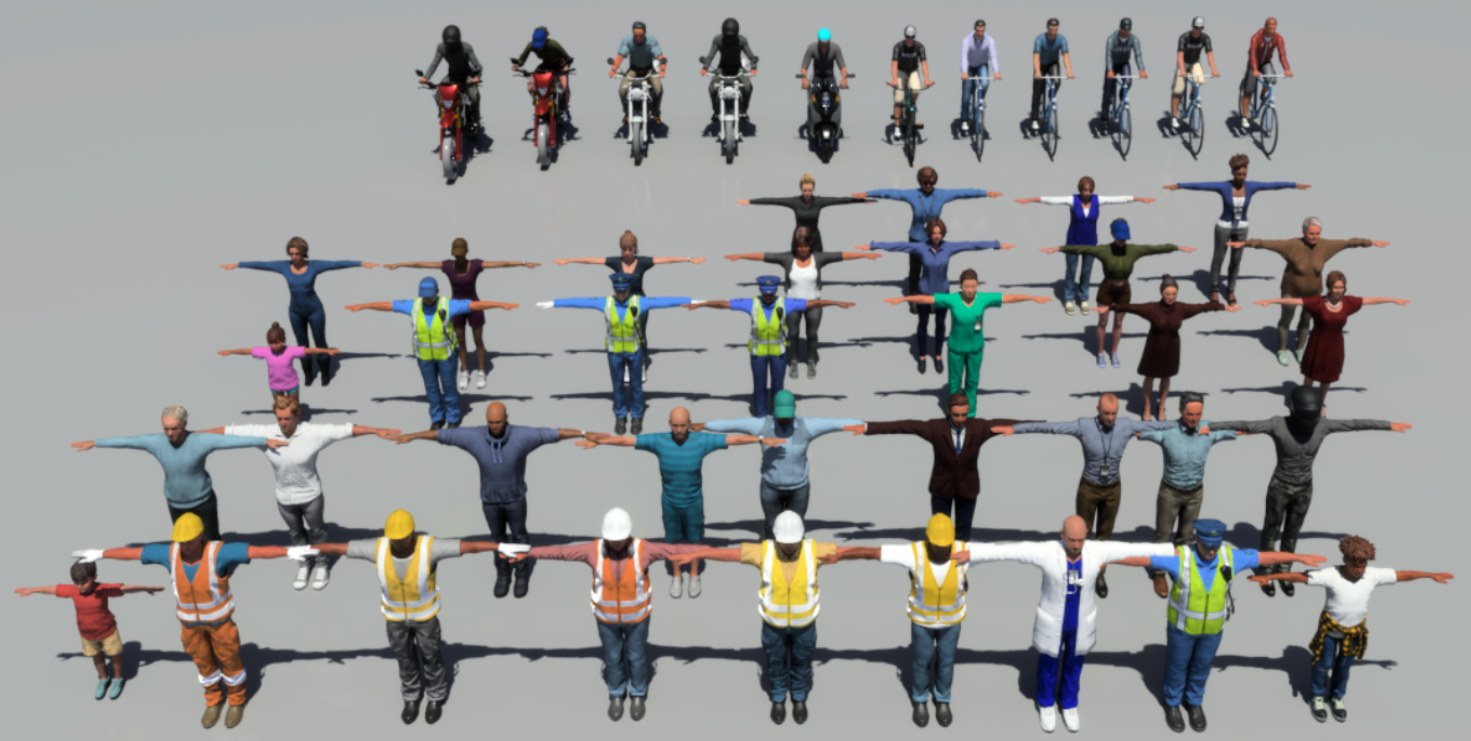}
\caption{The list of 3D synthetic assets used for improving obstacle detection targeting VRUs}
\label{fig:obstacle-asset-list}
\end{figure*}

\begin{figure*}
\centering
\includegraphics[width=0.6\linewidth]{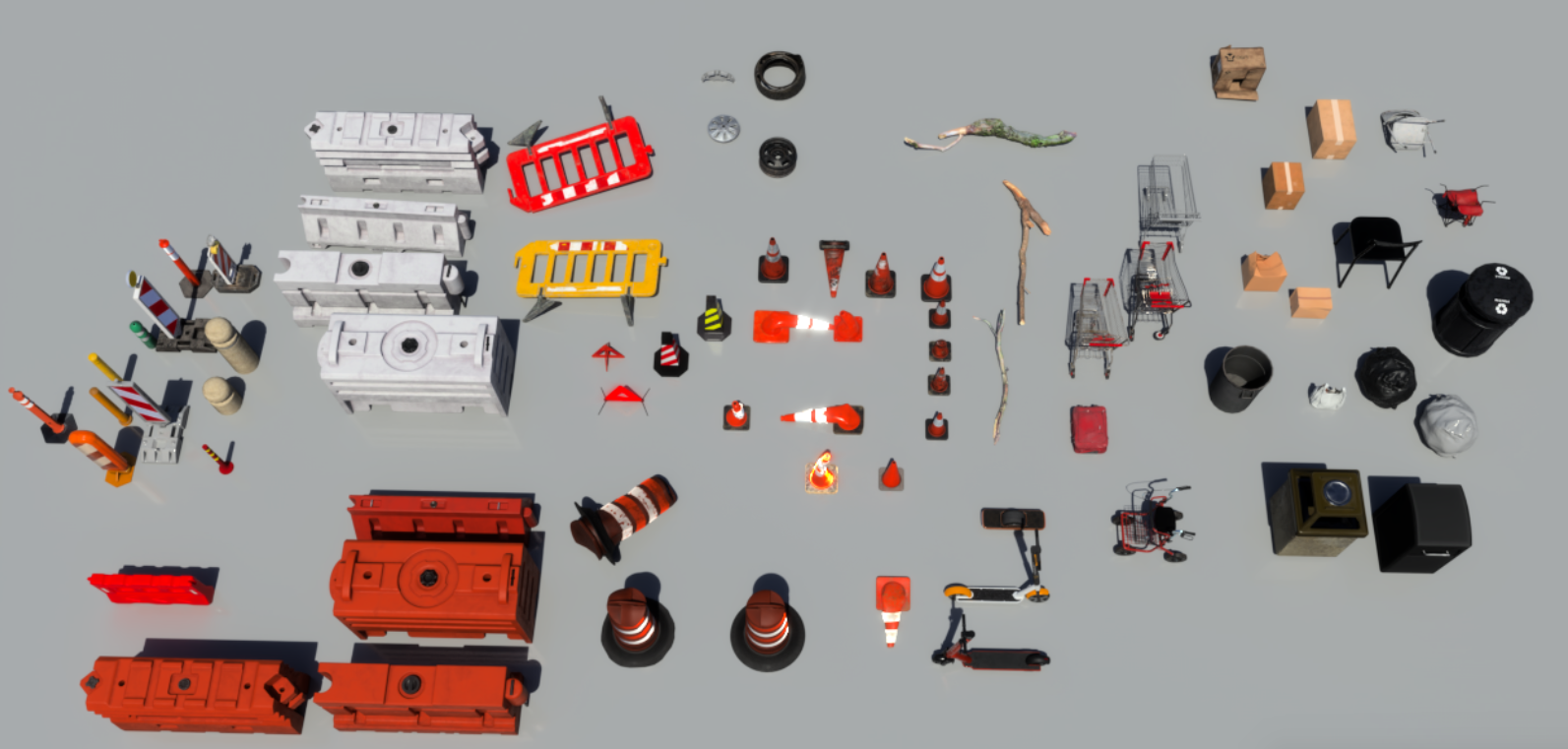}
\caption{The list of 3D synthetic assets used for improving freespace detection targeting closed-range hazards}
\label{fig:freespace-asset-list}
\end{figure*}

\begin{figure}[tb]
\centering
\includegraphics[width=0.8\linewidth]{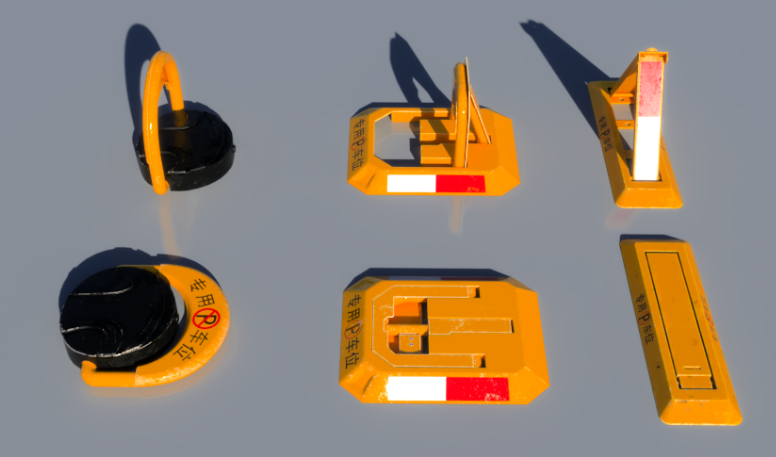}
\caption{The list of 3D synthetic assets used for improving parking lock detection}
\label{fig:groundlock-asset-list}
\end{figure}

\begin{figure*}
\centering
\includegraphics[width=1.0\linewidth]{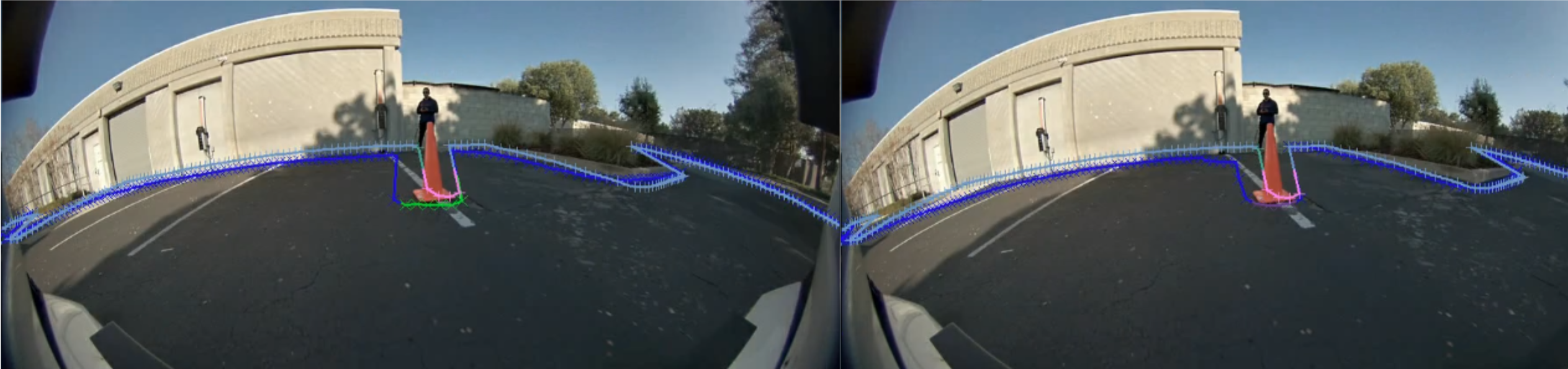}
\caption{Freespace prediction results on the real validation dataset. (Left) Inference with the model trained only on a real dataset, the traffic cone is mispredicted as a pedestrian. (Right) Freesapace prediction with real+ARSim with the same traffic cone correctly classified as hazard}
\label{fig:freespace-infer}
\end{figure*}

\section{Improving detection using ARSim data}
\label{sec:supp_arsim_improvement}
\subsection{3D Synthetic Assets} Figures \ref{fig:obstacle-asset-list}, \ref{fig:freespace-asset-list}, and \ref{fig:groundlock-asset-list} display the 3D synthetic assets utilized in generating data for obstacle detection, freespace detection, and parking detection tasks. In the obstacle detection data generation process, various VRUs are employed (Fig. \ref{fig:obstacle-asset-list}), each assigned random animations chosen from a pool of 10 different actions such as walking, jogging, bending, tying shoes, and picking up objects from the ground. These diverse animations, combined with the deployment of 36 distinct instances of 3D synthetic pedestrians, result in a wide array of variations within the dataset. To enhance freespace detection, we curated closed-range hazard data by incorporating a diverse array of 74 synthetic 3D assets distributed across 12 categories. These categories encompassed personal vehicles (such as electric scooters and onewheels), as well as various objects commonly encountered on roads, including trash cans, bags, shopping carts, tree branches, tires, boxes, chairs, traffic poles, movable barriers, traffic barriers, and traffic cones/triangles. Visual representations of these assets are depicted in Figure \ref{fig:freespace-asset-list}. Leveraging the frequency of occurrence of these categories on roads, we employed stratified sampling, as described earlier, to ensure a controlled distribution of assets in the generated ARSim data. For parknet ground lock detection improvement, we introduced three types of ground locks, each available in locked and unlocked states, as illustrated in Figure \ref{fig:groundlock-asset-list}. Unlike the placement strategy for obstacle and freespace detection data, the positioning of these ground locks differs. In the real-drive input data for parknet, parking spots are delineated by polygons in the ego coordinate system and assigned various attributes such as availability and existing ground lock presence. Utilizing these attributes, we deploy a ground lock within a parking spot if it remains unoccupied and lacks an existing ground lock.


\begin{figure*}
  \centering
  \begin{subfigure}{0.49\linewidth}
  \includegraphics[width=\linewidth]{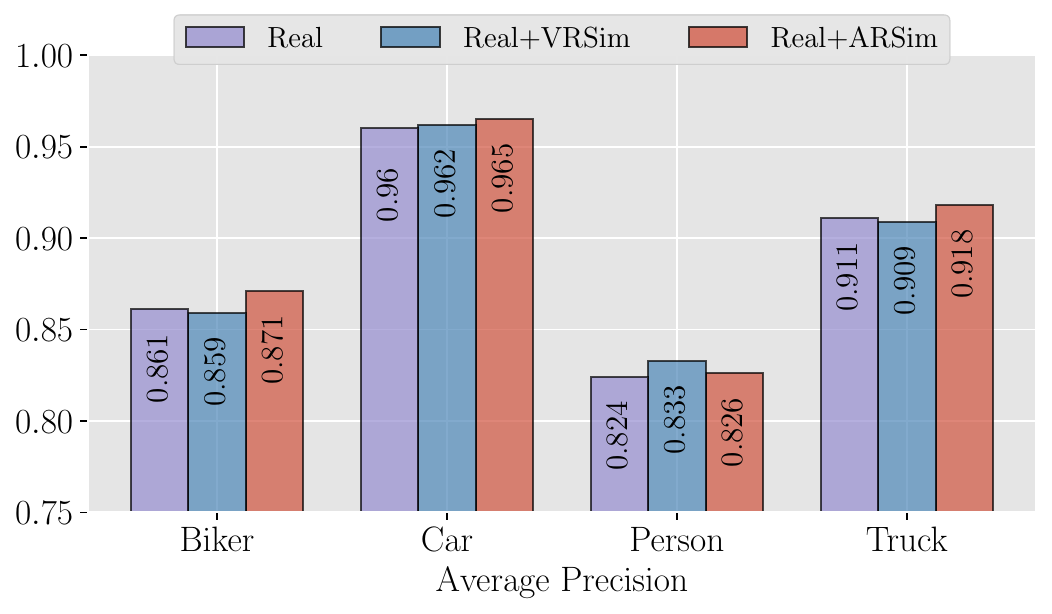}
    \label{fig:ablation-test-ap}
  \end{subfigure}
  \hfill
  \begin{subfigure}{0.49\linewidth}
  \includegraphics[width=\linewidth]{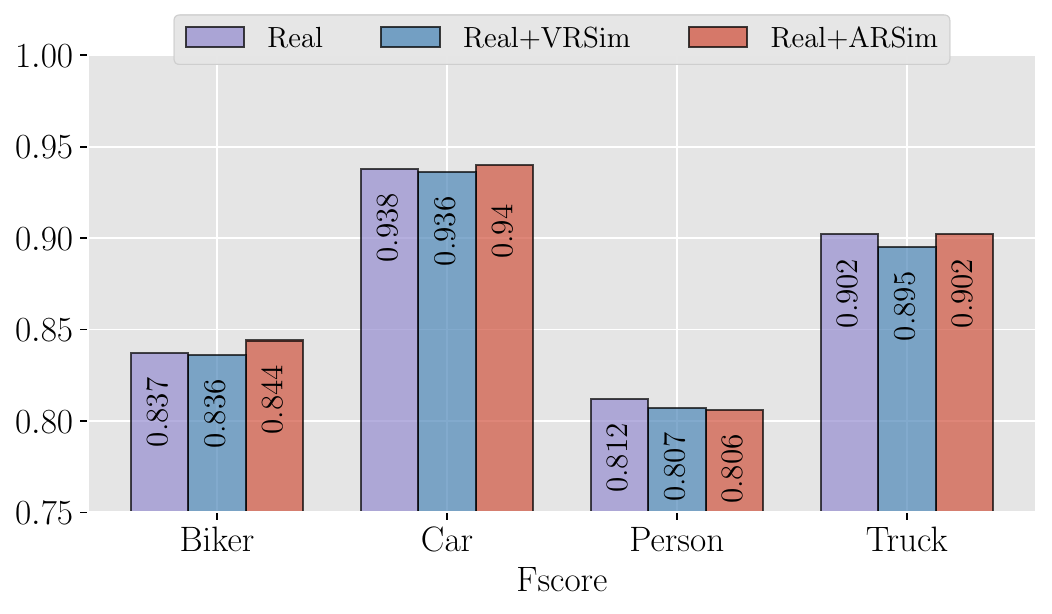}
    \label{fig:ablation-test-fscore}
  \end{subfigure}
  \caption{KPI evaluation on validation (\texttt{Real (val)}) data}
  \label{fig:ablation-test-result}
\end{figure*}

\subsection{Visualizing the improvement}
We visualize the freespace hazard detection results by inferring on a test scene the models trained with and without the ARSim data as explained in the main paper body. Fig. \ref{fig:freespace-infer} showcases an example of correctly classified hazard objects from the test data, which were previously misclassified as pedestrians when trained solely on real data.

\begin{table}[tb]

  \centering
\begin{tabular}{lcccc}
\toprule
Dataset     & Scenes &  Person & Biker \\
\midrule
\texttt{Real}           & 2.2M	                                & 	6.3M                & 	929K                           \\
\texttt{Real (val)}     & 306K                                & 266K                & 29K                           \\
\texttt{ARSim}             & 51K                            & 31K                & 17K                           \\
\texttt{VRSim}             & 51K                              & 31K                & 17K                           \\
\bottomrule   
\end{tabular}
  \caption{Statistics of the dataset used for ARSim/VRSim ablation test
  }
  \label{tab:ablation-data-supp}
\end{table}

\subsection{NCAP Ablation Study: Non-targeted case}

In the main body of the paper, we conducted ablation tests for obstacle3d detection specifically focusing on NCAP dummies. These tests involved comparing the performance of models trained using three different datasets: Real only, Real+VRSim, and Real+ARSim. Additionally, these trained models underwent evaluation on a separate generic obstacle3d validation dataset, referred to as \texttt{Real (target: NCAP Only)}. This validation set comprised classes such as biker (non-dummy real bikers), person (non-dummy real pedestrians), and truck, as detailed in Table \ref{tab:ablation-data-supp}. This validation step aimed to ensure that the inclusion of simulated data did not adversely affect performance metrics for non-targeted cases. Fig. \ref{fig:ablation-test-result} presents the average precision and F-score results, revealing that the differences in performance metrics were negligible. It's crucial to emphasize that Fig.~\ref{fig:ablation-test-result} was generated using existing data (used in the main body of the paper), and no new models were trained.


\end{document}